\documentclass[conference]{IEEEtran}
\IEEEoverridecommandlockouts

\usepackage{cite}
\usepackage{amsmath,amssymb,amsfonts}
\usepackage{graphicx}
\usepackage{booktabs}
\usepackage{multirow}
\usepackage{makecell}
\usepackage{array}
\usepackage{xcolor}
\usepackage{url}
\usepackage[hidelinks]{hyperref}
\usepackage{xspace}

\newcommand{\rctot}{RadOT-Eval\xspace}
\newcommand{\rexval}{ReXVal\xspace}
\newcommand{\radevalx}{RadEvalX\xspace}
\newcommand{\rexerr}{ReXErr-v1\xspace}
\newcommand{\green}{\mbox{GREEN-radllama2-7B}\xspace}
\newcommand{\sigerr}{clinically significant\xspace}
\newcommand{\insigerr}{clinically insignificant\xspace}

\begin{document}
\setlength{\textfloatsep}{6pt plus 1pt minus 2pt}
\setlength{\floatsep}{6pt plus 1pt minus 2pt}
\setlength{\intextsep}{6pt plus 1pt minus 2pt}

%\title{RadOT-Eval: Optimal Transport for Evaluating Radiology Report Generation via Structured Evidence}
\title{RadOT-Eval: Auditable Structured-Evidence Transport for Radiology Report Evaluation}

\author{
    \IEEEauthorblockN{
        Weixin Liu$^{1}$, 
        Juming Xiong$^{1}$, 
        Yang Li$^{1}$, 
        Qingyuan Song$^{1}$, \\
        Susannah Rose$^{2}$, 
        Murat Kantarcioglu$^{3}$, 
        Bradley Malin$^{1,2}$, 
        Zhijun Yin$^{1,2}$
    }
    \IEEEauthorblockA{$^{1}$Vanderbilt University, Nashville, TN, USA}
    \IEEEauthorblockA{$^{2}$Vanderbilt University Medical Center, Nashville, TN, USA}
    \IEEEauthorblockA{$^{3}$Virginia Tech, Blacksburg, VA, USA}
    \IEEEauthorblockA{
        Email: \{weixin.liu, juming.xiong, yang.li.2, qingyuan.song\}@vanderbilt.edu, \\
        muratk@vt.edu, 
        \{susannah.rose, b.malin, zhijun.yin\}@vumc.org
    }
}

\maketitle

\begin{abstract}
Automatic evaluation is critical for high-stakes text generation, where errors often involve omitted findings, hallucinated content, polarity reversals, location changes, uncertainty mismatches, and temporal-comparison errors rather than low surface similarity alone. Radiology report generation provides a challenging test case because generated reports must preserve structured clinical evidence across sources. We present \rctot, an interpretable structured-evidence optimal transport framework for offline auditing of radiology report generation. \rctot decomposes reference and candidate reports into attribute-structured clinical evidence units, aligns corresponding evidence using entropy-regularized optimal transport, and uses clinically meaningful side-channel discrepancies in a monotone risk model to predict error burden. All transport, feature, and readout choices are selected using the \rexval dataset, and the frozen system is evaluated on the independent \radevalx dataset. \rctot achieves Spearman correlations of $0.715$, $0.548$, and $0.399$ with total, \sigerr, and \insigerr annotated error burden, respectively, yielding higher point estimates than standard evaluation metrics and the open-source large language model (LLM)-based evaluator \green. In a frozen auxiliary corruption-sensitivity stress test on \rexerr, \rctot achieves $0.768$ AUROC and a $0.990$ corrupted-greater-than-clean paired win rate. These results show that structured evidence transport provides an auditable, rank-oriented evaluation tool for high-stakes generated clinical text under \rexval-only model selection and frozen \radevalx testing.
\end{abstract}

\begin{IEEEkeywords}
health informatics, radiology report evaluation, high-stakes text generation, optimal transport, interpretable model auditing
\end{IEEEkeywords}

\section{Introduction}

Automatic evaluation is a central problem for generated-text systems because surface overlap can miss factual inconsistency, unsupported content, and omissions \cite{maynez2020faithfulness,kryscinski2020factcc,pagnoni2021frank}. This issue is especially important in high-stakes domains, where generated documents must preserve local facts, negation, uncertainty, temporal context, and domain-specific risk. Radiology report generation is a representative setting because AI-generated reports may omit clinically important findings, hallucinate unsupported findings, reverse polarity, change anatomy or laterality, or misstate comparison with a prior study \cite{jing2018automatic,liu2019clinically,miura2021factual,yu2023evaluating}. Such errors can be clinically significant even when the candidate report has a high lexical or semantic overlap with the reference report \cite{papineni2002bleu,lin2004rouge,zhang2020bertscore,radevalx}.

Existing radiology report evaluation metrics leave a gap for auditable discrepancy prediction that can be assessed without tuning on the final evaluation benchmark. Lexical, learned, and embedding-based metrics provide useful global similarity signals, but they do not expose which clinical statements were aligned or which local attributes disagreed \cite{papineni2002bleu,lin2004rouge,sellam2020bleurt,zhao2019moverscore,rei2020comet,yuan2021bartscore}. While clinical label- and graph-based metrics such as CheXbert, RadGraph, RadGraph-XL, and RadCliQ incorporate domain structure, they could still obscure the local evidence behind a score \cite{smit2020chexbert,jain2021radgraph,jain2024radgraphxl,yu2023evaluating}. Large language model (LLM)-based and radiology-specific evaluators such as \green and RaTEScore can produce clinically oriented judgments or relevance scores. However, their outputs are not easily decomposed into stable local alignment decisions \cite{ostmeier2024green,zhao2024ratescore}. This limitation is especially important for cross-source evaluation, where expert-annotated radiology benchmarks are typically small. Methods that require target-set tuning after inspecting target performance may overfit to the evaluation set rather than measure true generalization.

\begin{figure*}[t]
\centering
\includegraphics[width=0.9\textwidth]{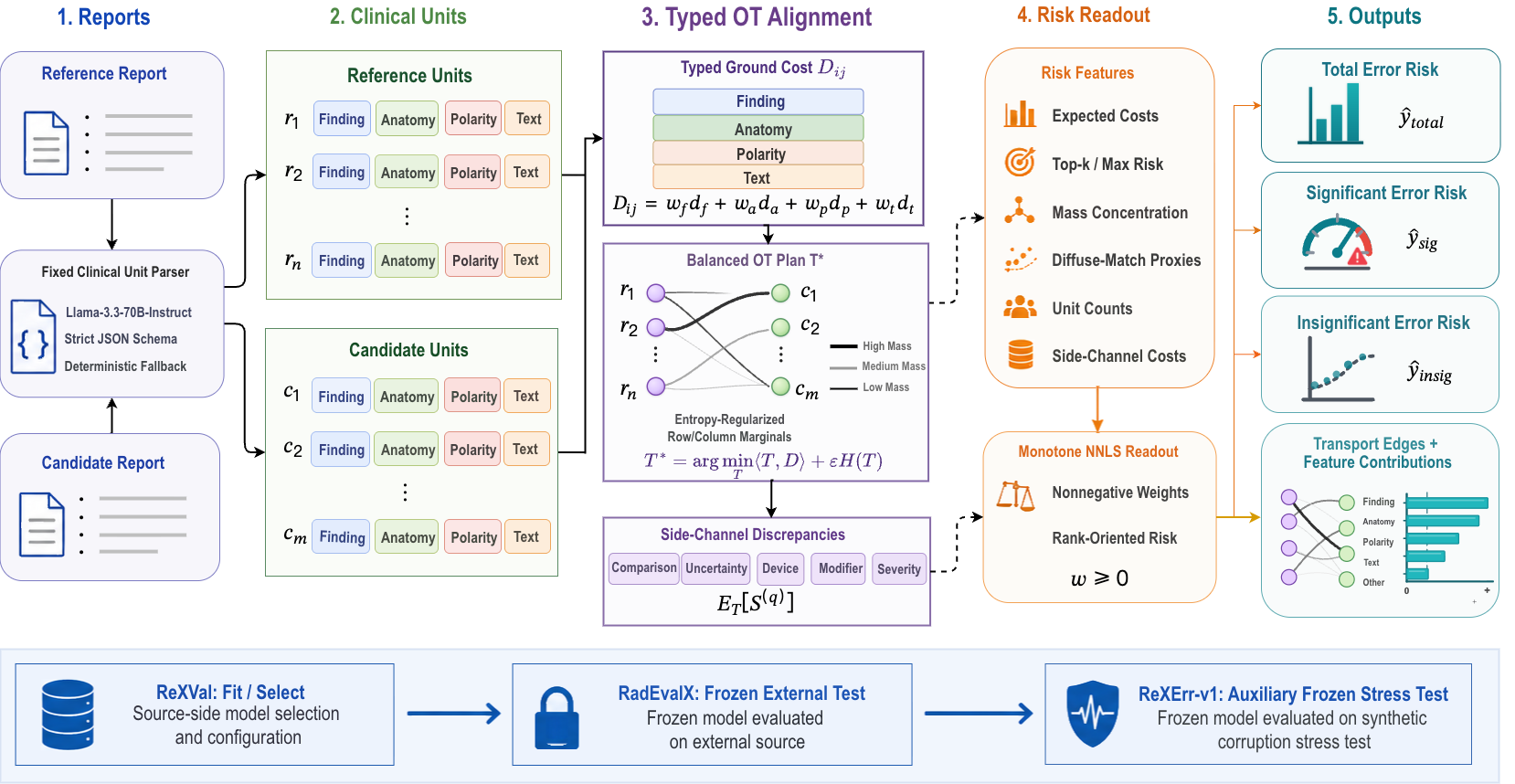}
\caption{Overview of \rctot. Reports are parsed into structured clinical units. Stable attributes define the OT ground cost and transport plan, while alignment-sensitive attributes are summarized as side-channel discrepancies. A monotone nonnegative readout combines alignment, discrepancy, concentration, diffuse-match, and unit-count features into rank-oriented error-risk scores.}
\label{fig:overview}
\end{figure*}

In this paper, we address this gap with \rctot (Fig.~\ref{fig:overview}), which formulates radiology report evaluation as structured clinical evidence transport. This formulation is motivated by the fact that corresponding clinical findings may differ in wording, ordering, and granularity, while clinically important errors often arise from localized changes in polarity, anatomy, uncertainty, severity, devices, or temporal comparison. \rctot decomposes each reference and candidate report into attribute-structured clinical units capturing findings, anatomy, polarity, uncertainty, comparison, device, modifier, severity, and supporting text. OT then aligns these units using relatively stable clinical anchors, including finding identity, anatomy, polarity, and lexical evidence. Unlike prior OT-based text metrics that transport over words or contextual token embeddings, \rctot transports over extracted clinical evidence units and separates evidence alignment from clinical risk readout. Under the learned transport plan, variation-sensitive attributes are measured as side-channel discrepancies. A monotone nonnegative readout combines OT-based alignment evidence and discrepancy signals into ranked risk scores for total, clinically significant, and clinically insignificant errors. This design yields rank-oriented risk scores and inspectable aligned evidence pairs, supporting offline model-development audit before prospective clinical evaluation.

We select transport, feature, and readout choices on the Radiology Report Expert Evaluation (\rexval) dataset \cite{rexval} and evaluate the frozen system on the independent Radiology Report Generation Models Evaluation Dataset for Chest X-rays (\radevalx) \cite{radevalx}. \rctot reaches Spearman correlations of $0.715$, $0.548$, and $0.399$ with total, \sigerr, and \insigerr error burden, yielding higher point estimates than the official \radevalx standard metric baselines and the open-source LLM-based evaluator \green. On the auxiliary synthetic-error benchmark \rexerr \cite{rao2024rexerr,rexerrdataset}, the frozen system achieves $0.768$ AUROC and a $0.990$ corrupted-greater-than-clean paired win rate. Ablations, shortcut controls, and robustness diagnostics show that high-severity errors are driven primarily by structured clinical discrepancies, whereas low-severity errors are more affected by report complexity. These analyses position \rctot as an applied, auditable evaluator for model-development workflows rather than a replacement for radiologist review.

% In summary, we have three contributions in this work. First, we formulate radiology report evaluation as structured evidence transport rather than scalar similarity scoring. Second, we separate alignment from risk readout: stable clinical attributes define the OT ground cost, while clinically sensitive attributes are scored as side-channel discrepancies under the transport plan. Third, we evaluate the evaluator under a source-separated protocol, selecting transport, feature, and readout choices on Radiology Report Expert Evaluation (\rexval) and testing once on RadEvalX (\radevalx), with auxiliary corruption-sensitivity and shortcut analyses used only diagnostically.

\section{Related Work}

\subsection{Structured evaluation of generated text}
Automatic text-generation evaluation has moved beyond surface overlap toward semantic, factual, and task-specific assessment. Lexical and consensus metrics such as BLEU, ROUGE, METEOR, CIDEr, and SPICE provide useful corpus-level similarity signals, but they do not directly verify local facts, negation, uncertainty, or temporal context \cite{papineni2002bleu,lin2004rouge,banerjee2005meteor,vedantam2015cider,anderson2016spice}. Embedding-based metrics, including BERTScore, BLEURT, MoverScore, COMET, BARTScore, and MAUVE, improve semantic matching but still return scalar scores without exposing aligned information units or structured attribute disagreements \cite{zhang2020bertscore,sellam2020bleurt,zhao2019moverscore,rei2020comet,yuan2021bartscore,pillutla2021mauve}. Factuality-oriented evaluation emphasizes unsupported content, omissions, and consistency errors, but local auditability remains limited \cite{maynez2020faithfulness,kryscinski2020factcc,pagnoni2021frank,scialom2021questeval}. \rctot follows this structured-evaluation view by comparing reference and candidate reports through explicit clinical units and their attributes, rather than through a single undifferentiated similarity score.

\subsection{Radiology report generation as a case study}
Radiology report generation has commonly been evaluated with metrics imported from machine translation, summarization, and image captioning \cite{jing2018automatic,wang2018tienet,chen2020m2trans,chen2021cmn}. These generic metrics are simple to compute but were not designed for clinical factuality, finding omission, hallucination, or temporal-comparison errors \cite{papineni2002bleu,lin2004rouge,banerjee2005meteor,miura2021factual}. Clinical label and graph metrics address part of this gap. CheXpert and CheXbert map reports to observation labels, while RadGraph and RadGraph-XL extract radiology entities and relations for graph-based comparison \cite{irvin2019chexpert,smit2020chexbert,jain2021radgraph,jain2024radgraphxl}. RadCliQ combines radiology-aware signals, including RadGraph-derived features, into a composite metric aligned with radiologist judgments \cite{yu2023evaluating}. These methods motivate the non-LLM baseline group in our experiments; specifically, we compare against the official \radevalx outputs for CheXbert, RadGraph-F1, and RadCliQ in Section~\ref{sec:baselines}.

\subsection{LLM-based radiology evaluators}
Recent work has explored generative or learned clinical evaluation for radiology reports. \green introduces generative radiology report evaluation and error notation using a radiology-tuned language model to produce clinically significant error explanations and a scalar evaluation score \cite{ostmeier2024green}. RaTEScore similarly reflects the trend toward radiology-specific learned evaluation rather than generic text similarity alone \cite{zhao2024ratescore}. These evaluators can provide clinically oriented judgments, but their decisions are not always decomposable into stable local alignment edges. Our goal is complementary: we evaluate whether structured evidence transport can achieve competitive cross-benchmark ranking while exposing unit-level alignments and monotone feature contributions.

\subsection{Optimal transport for structured text comparison}
Optimal transport provides a natural framework for comparing distributions or sets under a ground cost, with foundations in Earth-Mover-style distances and modern computational formulations \cite{rubner2000emd,villani2009optimal,santambrogio2015optimal,peyre2019ot}. Entropy regularization and Sinkhorn scaling make OT practical for repeated pairwise comparisons \cite{cuturi2013sinkhorn,altschuler2017sinkhorn}. Prior OT-based text metrics, such as Word Mover's Distance and MoverScore, transport mass over words or contextual token representations to produce document-level similarity scores \cite{kusner2015wmd,zhao2019moverscore}. \rctot differs in both support and objective: it transports mass over extracted clinical evidence units, uses a typed ground cost over stable clinical attributes, and then reads out clinically sensitive side-channel discrepancies under the resulting plan. Thus, OT is used not as a generic semantic distance, but as an auditable alignment layer for offline report-generation auditing. We also consider partial and unbalanced OT variants as ablations \cite{figalli2010partial,chizat2018unbalanced}.

\section{Problem and Data}

\subsection{Pair-level error prediction}
% In the general setting, each example consists of a reference document and a candidate document. 
In this radiology instantiation, each example consists of a reference report $R$ and a candidate report $C$. The goal is to predict three human-annotated aggregate targets: total errors, \sigerr errors, and \insigerr errors. We treat these outputs as rank-oriented error-risk scores rather than calibrated absolute error-count estimates.

\subsection{Benchmarks and preprocessing}
\rexval contains 200 report pairs from the Medical Information Mart for Intensive Care Chest X-ray (MIMIC-CXR) dataset, covering 50 studies with four retrieved candidate reports per study and radiologist error annotations \cite{rexval,johnson2019mimiccxr,goldberger2000physionet}. It is used only for source-side fitting and selection. \radevalx contains 100 Indiana University Chest X-ray (IU-Xray) reference reports paired with Memory-Driven Transformer (M2Tr)-generated candidates and expert error annotations; it is used only for final external evaluation \cite{radevalx,demner2016iuxray,chen2020m2trans,goldberger2000physionet}. The official \radevalx metric file provides per-pair scores for BLEU-4, BLEU-2, BERTScore, CheXbert, RadGraph-F1, and RadCliQ, enabling aligned baseline comparisons \cite{papineni2002bleu,zhang2020bertscore,smit2020chexbert,jain2021radgraph,yu2023evaluating}. Together, these resources define the benchmark setting for our applied evaluation study and underscore the limited scale of expert-labeled evaluation resources in the medical domain.

In addition, we prepare \rexerr as an auxiliary synthetic binary stress test based on clinically meaningful synthetic report errors \cite{rao2024rexerr,rexerrdataset}. We form 2,708 clean self-pairs and 2,708 error-injected pairs. Clean pairs receive zero synthetic error labels, while corrupted pairs contain sentence-level synthetic perturbations. We use \rexerr only to test whether the frozen evaluator assigns higher risk to corrupted reports than to clean self-pairs. We do not use its synthetic error counts or categories as fine-grained calibration targets because they are generated perturbation labels rather than radiologist-annotated report-level error burdens, and their severity distribution is not directly matched to the \rexval or \radevalx annotation schemes. Table~\ref{tab:data_preproc} summarizes benchmark roles and clinical-unit preprocessing.

\begin{table}[t]
\centering
\scriptsize
\caption{Benchmarks, roles, and clinical-unit preprocessing. Unit and fallback counts for \rexval and \radevalx are reported jointly; \rexerr is used only for frozen binary stress testing.}
\label{tab:data_preproc}
\resizebox{\columnwidth}{!}{%
\begin{tabular}{lccccc}
\toprule
Dataset & Pairs & Source & Role & Units & Fallback roles \\
\midrule
\rexval & 200 & MIMIC-CXR & Fit/select only & \multirow{2}{*}{1,960} & \multirow{2}{*}{7} \\
\radevalx & 100 & IU-Xray & External test only & & \\
\rexerr & 5,416 & MIMIC-CXR & Frozen binary stress test only & 66,342 & 2,265 \\
\bottomrule
\end{tabular}}
\end{table}

\subsection{Clinical-unit extraction}
\label{sec:clinical_unit_extraction}
Each report is converted into attribute-structured clinical units using a fixed \texttt{Llama-3.3-70B-Instruct} parser \cite{grattafiori2024llama}. The prompt requests strict JSON with span text, canonical finding, surface finding, polarity, uncertainty, comparison, device, severity, anatomy, modifiers, and confidence fields. The schema is summarized in Appendix~\ref{app:parser_schema}; released repaired unit tables allow downstream reproduction without rerunning the 70B parser.

% We use the same prompt, schema, normalization code, decoding procedure, and fallback rule for \rexval, \radevalx, and \rexerr. 
% The parser stage is a fixed information-extraction step: it does not train or update model parameters, and it is run deterministically rather than sampled. If the model response cannot be parsed as valid JSON, we retry the same structured extraction procedure up to three times. Benchmark labels, error counts, and target categories are never provided to the parser or used in canonicalization, fallback triggering, or report deduplication. If parsing fails or produces no clinical units for a non-empty report, punctuation-segmented sentences are used as fallback units, processed by the same rule-based canonicalizers, and assigned a fixed fallback confidence of $0.50$. For \rexerr, identical report texts are parsed once and reused across repeated report-role assignments only for computational efficiency. We store parser outputs, fallback indicators, unit tables, and run metadata so that the downstream transport, readout, and evaluation stages can be reproduced from released repaired unit tables without rerunning the 70B parser.

\section{RadOT-Eval: Structured Evidence Transport}

\rctot is a structured evidence-alignment framework for paired documents that can be decomposed into domain-specific information units. In radiology report evaluation, these units are clinical statements with attributes such as finding, anatomy, polarity, uncertainty, comparison, device, modifier, and severity. The key technical design is to use OT only for auditable evidence alignment: stable attributes define the transport ground cost, while clinically sensitive attributes are withheld from matching and scored later as side-channel discrepancies. The method has three stages: unit extraction, auditable transport alignment, and monotone risk readout.

\subsection{Attribute-structured clinical-unit alignment}

For a pair of reports, optimal transport represents alignment as a nonnegative matrix $T$, where $T_{ij}$ is the amount of mass assigned between reference unit $r_i$ and candidate unit $c_j$. The transport plan is optimized with respect to a ground-cost matrix $D$, whose entry $D_{ij}$ measures the mismatch between the two units. In \rctot, let $R=\{r_1,\ldots,r_n\}$ and $C=\{c_1,\ldots,c_m\}$ denote the extracted reference and candidate clinical units. For each pair $(r_i,c_j)$, we compute attribute-specific alignment costs for finding, anatomy, polarity, and text:
\begin{align}
D_{ij}={}&w_f d_{\mathrm{find}}(r_i,c_j)+w_a d_{\mathrm{anat}}(r_i,c_j) \nonumber\\
&+w_p d_{\mathrm{pol}}(r_i,c_j)+w_t d_{\mathrm{text}}(r_i,c_j).
\end{align}
The weights are nonnegative and sum to one, i.e., $w_f,w_a,w_p,w_t\geq 0$ and $w_f+w_a+w_p+w_t=1$. They control the relative emphasis placed on finding identity, anatomy, polarity, and lexical evidence when constructing the transport ground cost. In the method description, we treat these weights as predefined experimental design choices rather than learned pairwise matchers. The candidate weight priors and the final selected setting are specified in the source-only selection protocol in Section~\ref{sec:source_selection}.

Let $a_i=1/n$ and $b_j=1/m$. The primary \rctot system computes the entropy-regularized balanced transport plan \cite{cuturi2013sinkhorn,peyre2019ot}
\begin{equation}
T^\star=\arg\min_{T\in U(a,b)} \sum_{i,j}T_{ij}D_{ij}+\epsilon\sum_{i,j}T_{ij}(\log T_{ij}-1),
\end{equation}
where $U(a,b)$ enforces the row and column marginals. The entropy parameter $\epsilon$ controls alignment smoothness: smaller values produce sharper transport plans, while larger values distribute mass more diffusely across plausible matches. The candidate values and final selected value are reported with the source-only model-selection protocol in Section~\ref{sec:source_selection}.

\subsection{Attribute-specific cost components}
All attribute-specific component costs are normalized to $[0,1]$, where $0$ denotes agreement and larger values denote weaker alignment or stronger discrepancy. We use deterministic costs rather than learned pairwise matchers so that the transport plan remains auditable. Let $\tau(x)$ denote lower-cased alphanumeric tokenization of text field $x$, and let
\[
J(A,B)=
\begin{cases}
1, & A=B=\emptyset,\\
0, & \mathbf{1}[A=\emptyset]+\mathbf{1}[B=\emptyset]=1,\\
\frac{|A\cap B|}{|A\cup B|}, & \mathrm{otherwise}
\end{cases}
\]
be the Jaccard similarity between token or label sets. Let $\bot$ denote a missing categorical value and define
$m_\bot(a,b)=\mathbf{1}[a=\bot]+\mathbf{1}[b=\bot]$. For categorical fields, we use
\[
c_{\mathrm{cat}}(a,b;\mu,\eta)=
\begin{cases}
0, & a=b=\bot,\\
\mu, & m_\bot(a,b)=1,\\
0, & a=b,\\
\eta, & \mathrm{otherwise},
\end{cases}
\]
where $\mu$ is the missing-value penalty and $\eta$ is the mismatch penalty. For set-valued label fields, after splitting normalized parser outputs on pipe, semicolon, or comma delimiters, define
$m_\emptyset(A,B)=\mathbf{1}[A=\emptyset]+\mathbf{1}[B=\emptyset]$ and use
\[
c_{\mathrm{set}}(A,B;\mu)=
\begin{cases}
0, & A=B=\emptyset,\\
\mu, & m_\emptyset(A,B)=1,\\
1-J(A,B), & \mathrm{otherwise}.
\end{cases}
\]

Table~\ref{tab:typed_costs} gives the exact alignment costs used in the transport ground cost. The finding cost first checks whether both units have the same nonempty canonical finding. If so, the cost is $0$. Otherwise, it backs off to token overlap over the concatenation of canonical finding, surface finding, and unit span. This fallback makes the alignment less brittle when the parser normalizes semantically similar findings differently.

\begin{table}[t]
\centering
\scriptsize
\caption{Alignment costs in the \rctot ground cost. All costs are clipped to $[0,1]$.}
\label{tab:typed_costs}
\begin{tabular}{p{0.20\columnwidth}p{0.70\columnwidth}}
\toprule
Component & Definition \\
\midrule
Finding $d_{\mathrm{find}}$ 
& $0$ for exact nonempty canonical-finding match; otherwise $1-J(\tau(x_r),\tau(x_c))$, where $x$ concatenates canonical finding, surface finding, and span text. \\
Anatomy $d_{\mathrm{anat}}$ 
& $c_{\mathrm{set}}(A_r,A_c;0.6)$ over normalized anatomy-label sets. \\
Polarity $d_{\mathrm{pol}}$ 
& $c_{\mathrm{cat}}(p_r,p_c;0.5,1.0)$ over normalized polarity labels. \\
Text $d_{\mathrm{text}}$ 
& $1-J(\tau(s_r),\tau(s_c))$ over unit span texts. \\
\bottomrule
\end{tabular}
\end{table}

Side-channel costs use the same deterministic cost families but are not included in $D_{ij}$. Categorical side attributes, such as comparison, uncertainty, and device status, are measured with $c_{\mathrm{cat}}$; set-valued side attributes, such as modifier labels, are measured with $c_{\mathrm{set}}$; and severity is mapped to an ordinal scale before taking an absolute difference. These side-channel discrepancies are summarized under the transport plan for risk readout rather than used to determine the alignment itself. The exact side-channel constants and severity mapping used in the experiments are fixed before external evaluation and reported in the source-only experimental protocol in Section~\ref{sec:source_selection}. Their influence is audited in Section~\ref{sec:ablation_results} by removing severity and diffuse/concentration readout features.

This design avoids a common failure mode in structured report comparison. If every attribute is placed in the ground cost, the transport solver may avoid matching units that refer to the same clinical entity but disagree on an important side attribute. For example, a reference unit indicating that an opacity has improved and a candidate unit indicating that the same opacity is unchanged should still be aligned before the comparison discrepancy is scored. \rctot therefore uses stable attributes for alignment and reserves comparison, uncertainty, device, modifier, and severity for risk readout.

\subsection{Monotone risk readout}
The transport cost determines alignment, but not all clinically important attributes should determine matching. \rctot computes side-channel discrepancy matrices $S^{(q)}$ for comparison, uncertainty, device, modifier, and severity, then summarizes each by
\begin{equation}
    \mathbb{E}_{T^\star}[S^{(q)}]=\sum_{i,j}T^\star_{ij}S^{(q)}_{ij}.
\end{equation}
This design keeps alignment conservative while giving the readout access to clinically meaningful differences. Finding, anatomy, polarity, and text decide which units are comparable; comparison, uncertainty, device, modifier, and severity determine how much risk is associated with the aligned discrepancies.

The final feature vector contains expected alignment and side-channel costs, maximum and top-$k$ costs, top-$k$ mass-weighted risks, effective number of transport edges, top-mass concentration, transport entropy, diffuse-match proxies, and reference and candidate unit counts. Because the primary system uses balanced OT, all mass is transported; therefore, our ``soft unmatched'' features should not be interpreted as literal untransported mass. Instead, they measure whether the transported mass of a unit is concentrated on a single counterpart or diffusely spread across many weak matches. For each reference unit, define
\[
\rho_i=\frac{\max_j T^\star_{ij}}{a_i},
\]
and for each candidate unit define
\[
\gamma_j=\frac{\max_i T^\star_{ij}}{b_j}.
\]
Both quantities are clipped to $[0,1]$. We compute reference and candidate diffuse-match masses as
\[
M_R=\sum_i a_i(1-\rho_i), \qquad
M_C=\sum_j b_j(1-\gamma_j),
\]
along with total and asymmetric summaries and low-confidence fractions
\[
\frac{1}{n}\sum_i \mathbf{1}[\rho_i<\tau_{\rm diff}],
\qquad
\frac{1}{m}\sum_j \mathbf{1}[\gamma_j<\tau_{\rm diff}].
\]
We use $\tau_{\rm diff}=0.55$ as a fixed diffuse-match convention, chosen before external evaluation and not tuned on \radevalx.
These features act as soft omission and addition proxies under balanced transport: they are small when units have a dominant counterpart and larger when alignment mass is diffuse. The readout is monotone nonnegative least squares, so stronger discrepancy evidence cannot reduce predicted risk. This constraint also makes feature contributions interpretable after scaling.

\subsection{Transport-constraint variants}
The main system uses balanced OT. As ablations, we evaluate partial OT, which transports only a fraction of mass, and unbalanced OT, which relaxes marginal constraints with a mass-variation penalty \cite{figalli2010partial,chizat2018unbalanced}. In partial OT, the plan satisfies
\[
\begin{aligned}
T\mathbf{1} &\leq a, \\
T^\top\mathbf{1} &\leq b, \\
\sum_{i,j}T_{ij} &= \tau,\qquad 0<\tau\leq 1,
\end{aligned}
\]
where $\tau$ is the transported mass. This allows a small amount of difficult mass to remain unmatched.

For unbalanced OT, we replace hard marginal equality with Kullback--Leibler (KL)-penalized marginal deviation:
\[
\begin{aligned}
T^\star_{\mathrm{UOT}}
=
\arg\min_{T\geq 0}\;&
\sum_{i,j}T_{ij}D_{ij}
+\epsilon\sum_{i,j}T_{ij}(\log T_{ij}-1) \\
&+\lambda_m \operatorname{KL}(T\mathbf{1}\,\|\,a)
+\lambda_m \operatorname{KL}(T^\top\mathbf{1}\,\|\,b).
\end{aligned}
\]
Here the generalized KL divergence is
\[
\operatorname{KL}(u\,\|\,v)
=
\sum_i
\left(
u_i\log\frac{u_i}{v_i}-u_i+v_i
\right).
\]
The mass penalty $\lambda_m$ controls how strongly the transported row and column masses must match the original unit masses. These variants use the same feature-generation and source-only selection principles. They are not used to replace the primary system; instead, they test whether relaxing balanced transport changes the trade-off across error severities.

\section{Experimental Protocol}

\subsection{Source-only model selection}
\label{sec:source_selection}
All model-selection choices for \rctot are made using \rexval only, including the transport-weight prior, entropy parameter, and readout configuration. For each predefined configuration, we run 5-fold GroupKFold by study and select the final system by macro held-out Spearman across the three targets. The selected model is frozen and evaluated once on \radevalx, preventing target-benchmark label leakage.

The predefined alignment-prior grid varies the emphasis on finding identity, anatomy, polarity, and lexical evidence, including uniform, finding-heavy, anatomy-heavy, polarity-heavy, text-heavy, text-light, no-text, no-polarity, stable-prior, and parser-robust settings. The selected primary system uses the polarity-heavy ground-cost prior, $(w_f,w_a,w_p,w_t)=(0.250,0.200,0.400,0.150)$, and entropy parameter $\epsilon=0.20$. Side-channel constants are fixed before external evaluation rather than tuned on \radevalx: comparison and uncertainty use $c_{\mathrm{cat}}(\cdot,\cdot;0.35,1.0)$, device status uses $c_{\mathrm{cat}}(\cdot,\cdot;0.2,1.0)$, and modifier labels use $c_{\mathrm{set}}(\cdot,\cdot;0.4)$. Severity is mapped to an ordinal scale, with none/normal $=0$, mild $=0.33$, moderate $=0.66$, severe/marked $=1.0$, empty $=0$, and other nonempty unknown values $=0.5$; severity discrepancy is the absolute difference between the two mapped values.

Appendix~\ref{app:selection} reports the selected source-side configuration and additional full metrics.

\noindent\textbf{Reproducibility controls.}
To support reproduction without rerunning the 70B parser, our reproducibility package includes repaired unit tables, feature matrices, selected configuration, fitted coefficients, prediction files, bootstrap outputs, ablation outputs, and evaluation scripts, subject to source-dataset data-use restrictions.

\subsection{Baselines and statistics}
\label{sec:baselines}
The baselines cover standard automatic metrics, an open-source LLM-based evaluator, non-ensemble discrepancy models, and shortcut controls. For the official \radevalx metrics, we use the distributed per-pair scores for BLEU-4, BLEU-2, BERTScore, CheXbert, RadGraph-F1, and RadCliQ \cite{papineni2002bleu,zhang2020bertscore,smit2020chexbert,jain2021radgraph,yu2023evaluating}. Similarity scores are converted to risk by $1-\mathrm{score}$, while RadCliQ is already treated as an error-risk score.

For the LLM-based baseline, we use the Stanford AIMI GREEN model, \texttt{StanfordAIMI/GREEN-RadLlama2-7b}, through the released \texttt{green\_score.GREEN} inference wrapper with the model's default reference-candidate report comparison prompt and output schema \cite{ostmeier2024green}. This checkpoint is a radiology-report evaluator fine-tuned from \texttt{StanfordAIMI/RadLLaMA-7b}. We run GREEN on the same 100 \radevalx reference-candidate report pairs used for \rctot evaluation. We report two GREEN-derived signals because the released evaluator exposes both a scalar report-level GREEN score and structured error annotations. The scalar score is a quality-oriented score, so we convert it to an error-risk direction as $1-\mathrm{GREEN}$ and report this as GREEN risk. The native structured count is computed from GREEN's six clinically significant error categories and is reported for total and clinically significant targets. We do not report a native GREEN count for \insigerr errors because the released GREEN schema does not provide a stable \radevalx-style clinically insignificant error-count field.

The non-ensemble discrepancy baselines include text-only ridge, structured monotone, graph-derived monotone, and a graph neural network (GNN) discrepancy model. The shortcut controls compare full \rctot with a variant that removes the reference-unit count and with a reference-count-only model.

Spearman correlation is the primary metric because evaluators are often used to rank reports by clinical risk. Pearson, Kendall, MAE, and RMSE are reported in Appendix~\ref{app:full_metrics}. For \sigerr screening, we report AUROC, AUPRC, sensitivity at high-specificity operating points, and top-20\% enrichment. Paired \radevalx comparisons use bootstrap 95\% confidence intervals for Spearman differences between \rctot and each comparator. 
\section{Results}

\subsection{Main RadEvalX transfer performance}
The selected \rctot system obtains Spearman correlations of $0.715$, $0.548$, and $0.399$ for total, \sigerr, and \insigerr errors, respectively. Individual bootstrap confidence intervals are $[0.594,0.803]$, $[0.395,0.677]$, and $[0.224,0.559]$. Pearson, Kendall, and uncalibrated score-scale MAE/RMSE metrics are reported in Appendix~\ref{app:full_metrics}; Spearman correlation is the primary endpoint. Table~\ref{tab:main} compares \rctot with official standard metrics, the open-source LLM-based evaluator baseline \green, non-ensemble discrepancy baselines, and shortcut controls.

\begin{table}[t]
\centering
\scriptsize
\caption{\radevalx Spearman correlations. Official standard metrics use distributed per-pair scores. GREEN risk denotes $1-\mathrm{GREEN}$ from the released scalar GREEN score, while GREEN native count denotes the sum of GREEN's native clinically significant error categories. Bold indicates the best non-diagnostic evaluator in each column, with ties bolded; the reference-count-only row is a shortcut diagnostic, not a clinical discrepancy baseline.}
\label{tab:main}
\begin{tabular}{lccc}
\toprule
System & Total & Sig. & Insig. \\
\midrule
BLEU-4 & 0.108 & 0.096 & 0.046 \\
BLEU-2 & 0.292 & 0.162 & 0.157 \\
BERTScore & 0.349 & 0.195 & 0.151 \\
CheXbert & 0.492 & 0.413 & 0.103 \\
RadGraph-F1 & 0.285 & 0.159 & 0.147 \\
RadCliQ & 0.335 & 0.188 & 0.177 \\
\midrule
GREEN native count & 0.296 & 0.382 & -- \\
GREEN risk & 0.570 & 0.433 & 0.141 \\
\midrule
Text-only ridge & 0.592 & 0.337 & 0.324 \\
Structured monotone & 0.612 & 0.398 & 0.256 \\
Graph-derived monotone & 0.569 & 0.469 & 0.206 \\
GNN discrepancy model & 0.607 & 0.398 & 0.119 \\
\midrule
Reference count only & 0.474 & 0.201 & 0.410 \\
\rctot w/o ref count & 0.712 & \textbf{0.548} & 0.308 \\
\textbf{\rctot} & \textbf{0.715} & \textbf{0.548} & \textbf{0.399} \\
\bottomrule
\end{tabular}
\end{table}

Among official standard metrics, CheXbert is strongest for total and \sigerr errors, while RadCliQ is strongest for \insigerr errors. \rctot improves the corresponding point estimates by $0.224$, $0.135$, and $0.222$ Spearman. Compared with GREEN risk, \rctot increases total, \sigerr, and \insigerr Spearman point estimates by $0.146$, $0.115$, and $0.258$. Paired bootstrap intervals support reliable gains over several official standard metrics for total and \sigerr errors, while gains over the strongest baselines remain more uncertain on the 100-pair \radevalx set. Detailed paired intervals are reported in Appendix~\ref{app:paired_tests}.

For \sigerr screening, \rctot improves AUROC/AUPRC from $0.696/0.745$ for GREEN risk to $0.766/0.826$, and improves sensitivity at 90\% specificity from $0.161$ to $0.403$ (upper panel of Table~\ref{tab:binary_eval}).

\subsection{Auxiliary ReXErr-v1 synthetic stress testing}
Before applying \rctot to \rexerr, we verify that the refit \rexval readout exactly reproduces the selected \radevalx predictions, with maximum absolute prediction differences below $10^{-14}$. We then use \rexerr only as a frozen auxiliary binary corruption-sensitivity stress test. This experiment asks whether the selected evaluator assigns higher risk to synthetic error-injected reports than to clean self-pairs from the same source distribution; it is not used for model selection, hyperparameter tuning, or fine-grained error-type calibration.

The lower panel of Table~\ref{tab:binary_eval} shows the stress-test summary. A reference-count-only control obtains chance performance, indicating that the clean-versus-corrupted distinction is not explained by report complexity alone. Full \rctot separates clean from corrupted pairs with an AUROC of $0.768$ and an AUPRC of $0.714$. When each corrupted pair is compared with its clean counterpart from the same source report, \rctot assigns higher risk to the corrupted report in $99.0\%$ of pairs. These results provide large-scale evidence that the frozen evaluator is sensitive to injected corruption, while the primary human-annotated evidence comes from \rexval-only selection followed by frozen evaluation on the independent \radevalx benchmark.

\begin{table}[t]
\centering
\scriptsize
\caption{Auxiliary binary evaluation. The upper panel screens for any \sigerr error on \radevalx; the lower panel tests frozen clean-versus-corrupted sensitivity on \rexerr. Sens.@80/90 denote sensitivity at 80\%/90\% specificity; Extra reports top-20\% enrichment for \radevalx and paired win rate for \rexerr.}
\label{tab:binary_eval}
\resizebox{\columnwidth}{!}{%
\begin{tabular}{llccccc}
\toprule
Task & System & AUROC & AUPRC & Sens.@80 & Sens.@90 & Extra \\
\midrule
\multirow{3}{*}{\makecell[l]{\radevalx\\sig. screen}}
& GREEN native count & 0.654 & 0.721 & 0.129 & 0.129 & -- \\
& GREEN risk & 0.696 & 0.745 & 0.452 & 0.161 & 1.048 \\
& \textbf{\rctot} & \textbf{0.766} & \textbf{0.826} & \textbf{0.613} & \textbf{0.403} & \textbf{1.452} \\
\midrule
\multirow{2}{*}{\makecell[l]{\rexerr\\corruption}}
& Reference count only & 0.500 & 0.500 & -- & -- & -- \\
& \textbf{\rctot} & \textbf{0.768} & \textbf{0.714} & -- & -- & \textbf{0.990} \\
\bottomrule
\end{tabular}}
\end{table}

\subsection{Ablation studies}
\label{sec:ablation_results}
Table~\ref{tab:ablation} summarizes the architecture and transport-constraint ablations. Mean transport cost alone is weak, indicating that report-level error risk is not captured by a single average alignment distance. Adding expected side-channel costs substantially improves total and \sigerr prediction. Top-$k$ and maximum-risk features further improve high-severity rank. Within the feature-ablation block, the full feature set gives the largest gain for \insigerr errors, largely through unit-count and mass-distribution signals.

The transport-constraint ablation shows a severity-specific trade-off. Partial OT improves \insigerr ranking to $0.453$, while the primary balanced \rctot remains strongest for total and \sigerr error prediction. Unbalanced OT improves total ranking among alternative variants but reduces \sigerr transfer.

\begin{table}[t]
\centering
\scriptsize
\caption{Architecture and transport-constraint ablations. Variants are selected on \rexval and evaluated once on \radevalx. Bold marks the best reported ablation value; the primary \rctot system is retained for main comparisons because it is strongest for total and \sigerr errors.}
\label{tab:ablation}
\begin{tabular}{lcccc}
\toprule
Variant & Macro & Total & Sig. & Insig. \\
\midrule
Mean transport only & 0.226 & 0.339 & 0.237 & 0.103 \\
+ side-channel expected & 0.398 & 0.561 & 0.479 & 0.154 \\
+ top-$k$/max & 0.429 & 0.629 & 0.518 & 0.141 \\
+ learned cost weights & 0.427 & 0.642 & 0.504 & 0.135 \\
\textbf{Full \rctot} & \textbf{0.554} & \textbf{0.715} & \textbf{0.548} & 0.399 \\
\midrule
Partial OT variant & 0.519 & 0.641 & 0.464 & \textbf{0.453} \\
Balanced OT variant & 0.509 & 0.637 & 0.469 & 0.422 \\
Unbalanced OT variant & 0.504 & 0.684 & 0.406 & 0.422 \\
\bottomrule
\end{tabular}
\end{table}

All sensitivity variants in Table~\ref{tab:sensitivity} are selected or audited using \rexval only and then evaluated frozen on \radevalx. Performance is stable across nearby entropy values, transport-weight priors, and readout-feature audits. Although some nearby settings have slightly higher \radevalx point estimates, they are not selected because the final configuration is chosen only by source-side \rexval GroupKFold performance. Removing severity features or diffuse/concentration features also preserves the main conclusions, suggesting that the reported transfer performance is not driven by a single hand-specified severity mapping or diffuse-match threshold.

\begin{table}[t]
\centering
\scriptsize
\caption{Sensitivity analysis for transport and readout choices. The bold row is the \rexval-selected primary system, not the highest \radevalx point estimate. Feat. denotes the number of readout features; Sig./Insig. denotes \sigerr/\insigerr Spearman correlations.}
\label{tab:sensitivity}
\resizebox{\columnwidth}{!}{%
\begin{tabular}{lcccccc}
\toprule
Variant & $\epsilon$ & Feat. & ReXVal & RadEvalX & Total & Sig./Insig. \\
\midrule
\textbf{Selected full model} & 0.20 & 66 & 0.846 & 0.554 & 0.715 & 0.548 / 0.399 \\
$\epsilon=0.05$ & 0.05 & 66 & 0.846 & 0.563 & 0.726 & 0.542 / 0.423 \\
$\epsilon=0.10$ & 0.10 & 66 & 0.844 & 0.565 & 0.734 & 0.558 / 0.404 \\
Uniform weights & 0.20 & 66 & 0.845 & 0.554 & 0.708 & 0.548 / 0.405 \\
Text-light weights & 0.20 & 66 & 0.839 & 0.562 & 0.722 & 0.565 / 0.400 \\
No text cost & 0.20 & 66 & 0.838 & 0.561 & 0.724 & 0.555 / 0.404 \\
No polarity cost & 0.20 & 66 & 0.839 & 0.546 & 0.709 & 0.535 / 0.396 \\
No severity features & 0.20 & 61 & 0.851 & 0.551 & 0.710 & 0.556 / 0.387 \\
No diffuse features & 0.20 & 49 & 0.845 & 0.558 & 0.709 & 0.556 / 0.409 \\
\bottomrule
\end{tabular}}
\end{table}

\subsection{Shortcut, robustness, and interpretability analyses}
Because the feature-importance analysis assigns high weight to reference-unit count for \insigerr errors, we perform shortcut audits. Table~\ref{tab:diagnostics} shows that total and \sigerr prediction are not explained by reference complexity: removing reference-unit count preserves total and \sigerr correlations at $0.712$ and $0.548$, while reference-count only is much weaker. By contrast, \insigerr errors are strongly complexity-driven: reference-count only reaches $0.410$, and the partial correlation after controlling for reference count drops to $0.077$. This pattern is an important severity-specific property of the target rather than only a model artifact.

Parser robustness is stable. Removing all three \radevalx pairs affected by sentence fallback preserves the main conclusions: the total, \sigerr, and \insigerr Spearman correlations become $0.708$, $0.535$, and $0.391$, respectively. Under random clinical-unit dropout, performance degrades smoothly; at 10\% unit dropout, the total and \sigerr Spearman correlations remain $0.674$ and $0.516$, and at 25\% dropout they remain $0.588$ and $0.469$.

\begin{table}[t]
\centering
\scriptsize
\caption{Shortcut and robustness diagnostics. Partial correlations are rank residual correlations after controlling for reference-unit count.}
\label{tab:diagnostics}
\begin{tabular}{lccc}
\toprule
Diagnostic & Total & Sig. & Insig. \\
\midrule
Full \rctot & 0.715 & 0.548 & 0.399 \\
\rctot without ref count & 0.712 & 0.548 & 0.308 \\
Reference count only & 0.474 & 0.201 & 0.410 \\
Partial corr. controlling ref count & 0.613 & 0.529 & 0.077 \\
Without fallback pairs & 0.708 & 0.535 & 0.391 \\
10\% unit dropout & 0.674 & 0.516 & 0.349 \\
25\% unit dropout & 0.588 & 0.469 & 0.264 \\
\bottomrule
\end{tabular}
\end{table}

Error-type analysis supports semantic interpretability (Table~\ref{tab:errtype}). Significant finding omissions are associated with side-channel maximum risk, top-risk summaries, and side-channel expected sums, with Spearman values up to $0.542$. Significant comparison omissions align with comparison expected cost, with Spearman $0.501$. Uncertainty omissions align with uncertainty features, although this category is rare in \radevalx and should be interpreted cautiously. Risk-bucket analysis also shows monotonic stratification: from the lowest to highest predicted quintile, mean true total errors increase from $0.70$ to $5.05$, \sigerr errors from $0.20$ to $2.85$, and \insigerr errors from $0.45$ to $1.90$.

\begin{table}[t]
\centering
\scriptsize
\caption{Representative \radevalx error-type alignment. Low-count categories are exploratory.}
\label{tab:errtype}
\begin{tabular}{llcc}
\toprule
Severity / category & Best aligned feature & Nonzero & Spearman \\
\midrule
Sig. finding omission & Side maximum sum & 56 & 0.542 \\
Sig. finding omission & Top-risk sum & 56 & 0.525 \\
Sig. comparison omission & Comparison expected & 14 & 0.501 \\
Sig. uncertainty omission & Uncertainty expected & 2 & 0.433 \\
Insig. finding omission & Reference unit count & 43 & 0.438 \\
Insig. comparison omission & Comparison top-3 mean & 7 & 0.317 \\
\bottomrule
\end{tabular}
\end{table}

Qualitative transport reconstructions show that high \sigerr cases are driven by edges involving missed opacity, pleural-effusion, comparison, or modifier content, while high \insigerr cases show multiple moderate mismatches and stronger unit-count or mass-distribution contributions. Representative transport-edge and feature-contribution tables are provided in Appendix~\ref{app:qualitative}.

\section{Discussion and Limitations}

The results support treating high-stakes generated-text evaluation as structured clinical-unit comparison rather than only scalar similarity estimation. \rctot uses OT as an auditable alignment layer, not as a claim of a new OT solver. The technical novelty is the domain-specific OT formulation: transport is performed over attribute-structured clinical units, the ground cost is restricted to stable alignment anchors, and clinically sensitive attributes are scored only after alignment as side-channel risk evidence. Because all choices are selected on \rexval and frozen before \radevalx evaluation, the transfer results are less likely to reflect target-benchmark tuning. \rctot yields higher point estimates than the official \radevalx metrics and \green on all three ranking targets, although intervals against the strongest baselines remain uncertain on the 100-pair external set. The readout should be interpreted as a rank-oriented risk score, not a calibrated error-count estimator.

The ablations clarify why structured evidence transport is useful. Mean transport cost alone is weak, indicating that report-level risk is not captured by a single average alignment distance. Side-channel discrepancies and top-$k$ risk summaries are important for high-severity errors, while unit-count and mass-distribution features explain part of the low-severity target. This supports the design choice to separate stable alignment dimensions from risk readout channels: findings, anatomy, polarity, and lexical evidence define which units are comparable, while comparison, uncertainty, device, modifier, and severity discrepancies are scored under the resulting transport plan. The analyses also reveal a severity-specific structure. Total and \sigerr prediction remain strong after controlling for reference complexity, whereas \insigerr prediction is more strongly associated with report complexity and cumulative mild mismatch burden.

The \rexerr experiment should be interpreted only as an auxiliary binary corruption-sensitivity stress test, not as a fine-grained external validation benchmark. It shows that the frozen evaluator is sensitive to injected corruptions in paired source-controlled comparisons, but the primary external evidence remains the human-annotated \radevalx benchmark. Several limitations remain. The main external benchmark contains only 100 pairs, which widens paired uncertainty intervals against strong baselines. \rctot also depends on clinical-unit extraction from a large local instruction model, making raw-text re-extraction more expensive than the downstream transport and readout stages. The official \green comparison uses the released six-category schema, which does not provide a stable \radevalx-style clinically insignificant count field. Finally, this paper studies a single non-ensemble evaluator without target-domain adaptation or joint parser training. Future work should study larger external benchmarks, improved anatomical and laterality normalization, calibrated count prediction, and sentence-level error localization. \rctot should be used as a rank-oriented audit signal for model development and targeted error analysis, not as an autonomous clinical decision tool.

\section{Conclusion}

We presented \rctot, a structured evidence-transport framework for auditable radiology report evaluation. The problem is to predict clinically meaningful report-level discrepancies that may be missed by scalar lexical, semantic, or label-overlap metrics. \rctot addresses this problem by decomposing reference and candidate reports into attribute-structured clinical units, aligning stable evidence with entropy-regularized OT, and reading out clinically relevant side-channel discrepancies through a monotone risk model. Under \rexval-only selection and frozen \radevalx evaluation, \rctot achieves strong ranking performance for total and clinically significant errors and remains competitive with official standard metrics and the open-source LLM-based evaluator baseline. The results suggest that structured evidence transport is a practical audit layer for model-development workflows in high-stakes generated clinical text, provided that model-selection choices are fixed before external testing.

\section*{Acknowledgment}
This research was, in part, funded by the Advanced Research Projects Agency for Health (ARPA-H). The views and conclusions contained in this document are those of the authors and should not be interpreted as representing the official policies, either expressed or implied, of the United States Government.

\appendices
\section{Additional Results}
\label{app:supplementary}

\subsection{Parser Schema and Category Definitions}
\label{app:parser_schema}

\begin{table}[tbp]
\centering
\scriptsize
\caption{Parser schema and \radevalx error-category definitions.}
\label{tab:schema_cat_app}
\resizebox{\columnwidth}{!}{%
\begin{tabular}{lp{0.62\columnwidth}}
\toprule
\multicolumn{2}{l}{\textbf{Clinical-unit schema}} \\
\midrule
\texttt{span\_text} & Verbatim report span supporting the unit \\
\texttt{canonical\_finding} & Normalized finding label for alignment \\
\texttt{surface\_finding} & Finding phrase as expressed in text \\
\texttt{polarity} & Assertion status, e.g., present, absent, uncertain \\
\texttt{uncertainty} & Certainty status, e.g., definite, probable, possible \\
\texttt{comparison} & Temporal comparison, e.g., stable, improved, worsened \\
\texttt{device} & Canonical device label or null \\
\texttt{attributes.severity} & Free-text severity normalized downstream \\
\texttt{anatomy} & Canonical anatomy labels with surface phrases \\
\texttt{modifiers} & Canonical modifier labels with surface phrases \\
\texttt{confidence} & Parser confidence; fallback units use $0.50$ \\
\midrule
\multicolumn{2}{l}{\textbf{\radevalx error categories}} \\
\midrule
1 & False prediction of finding \\
2 & Omission of finding \\
3 & Incorrect location or position of finding \\
4 & Incorrect severity of finding \\
5 & Unsupported mention of comparison \\
6 & Omission of comparison with previous study \\
7 & Unsupported mention of uncertainty \\
8 & Omission of uncertainty present in reference \\
\bottomrule
\end{tabular}}
\end{table}

\subsection{Source-Side Selection and Full Metrics}
\label{app:selection}
\label{app:full_metrics}

\begin{table}[tbp]
\centering
\scriptsize
\caption{Source-side selection and full \radevalx metrics. Bold indicates the selected source-side configuration.}
\label{tab:selection_fullmetrics_app}
\resizebox{\columnwidth}{!}{%
\begin{tabular}{lcccc}
\toprule
\multicolumn{5}{l}{\textbf{\rexval GroupKFold selection}} \\
\midrule
Ground cost & $\epsilon$ & Macro & Total & Sig./Insig. \\
\midrule
\textbf{Polarity-heavy} & \textbf{0.20} & \textbf{0.846} & \textbf{0.900} & \textbf{0.842 / 0.797} \\
Text-heavy & 0.20 & 0.846 & 0.897 & 0.846 / 0.794 \\
Polarity-heavy & 0.05 & 0.846 & 0.899 & 0.844 / 0.794 \\
Uniform-basic & 0.20 & 0.845 & 0.899 & 0.844 / 0.792 \\
Finding-heavy & 0.05 & 0.845 & 0.898 & 0.844 / 0.793 \\
\midrule
\multicolumn{5}{l}{\textbf{Full \radevalx metrics for selected \rctot}} \\
\midrule
Target & MAE & RMSE & Pearson & Spearman / Kendall \\
\midrule
Total & 11.692 & 12.122 & 0.713 & 0.715 / 0.556 \\
Significant & 6.786 & 7.350 & 0.524 & 0.548 / 0.421 \\
Insignificant & 5.085 & 5.274 & 0.439 & 0.399 / 0.310 \\
\bottomrule
\end{tabular}}
\end{table}

\subsection{Paired Statistical Tests}
\label{app:paired_tests}

\begin{table}[tbp]
\centering
\scriptsize
\caption{Paired bootstrap confidence intervals against official \radevalx standard metrics. $\Delta\rho$ is \rctot minus comparator.}
\label{tab:paired_full_app}
\resizebox{\columnwidth}{!}{%
\begin{tabular}{llccc}
\toprule
Target & Comparator & $\rho_c$ & $\Delta\rho$ & 95\% CI \\
\midrule
Total & BERTScore & 0.349 & 0.366 & [0.179,0.556] \\
Total & BLEU-2 & 0.292 & 0.424 & [0.224,0.624] \\
Total & BLEU-4 & 0.108 & 0.608 & [0.370,0.833] \\
Total & CheXbert & 0.492 & 0.224 & [0.052,0.406] \\
Total & RadCliQ & 0.335 & 0.381 & [0.203,0.567] \\
Total & RadGraph-F1 & 0.285 & 0.431 & [0.241,0.617] \\
\midrule
Sig. & BERTScore & 0.195 & 0.353 & [0.151,0.549] \\
Sig. & BLEU-2 & 0.162 & 0.386 & [0.169,0.608] \\
Sig. & BLEU-4 & 0.096 & 0.452 & [0.213,0.677] \\
Sig. & CheXbert & 0.413 & 0.135 & [-0.064,0.326] \\
Sig. & RadCliQ & 0.188 & 0.360 & [0.162,0.548] \\
Sig. & RadGraph-F1 & 0.159 & 0.389 & [0.187,0.600] \\
\midrule
Insig. & BLEU-4 & 0.046 & 0.352 & [0.109,0.593] \\
Insig. & CheXbert & 0.103 & 0.296 & [0.076,0.523] \\
Insig. & RadCliQ & 0.177 & 0.222 & [0.031,0.409] \\
\bottomrule
\end{tabular}}
\end{table}

\subsection{Qualitative Reconstructions}
\label{app:qualitative}

\begin{table*}[t]
\centering
\scriptsize
\caption{Representative high-risk transport edges from qualitative reconstructions. MW denotes mass-weighted; edges are selected for explanation only and are not used for model selection.}
\label{tab:edges_app}
\resizebox{\textwidth}{!}{%
\begin{tabular}{llp{0.30\textwidth}p{0.30\textwidth}cccc}
\toprule
Case & Rank & Reference unit & Candidate unit & Mass & Align & Side & MW risk \\
\midrule
report699 & 1 & opacity on the lateral view over the heart, present on previous exam, suggesting chronic subsegmental atelectasis or scarring & heart and cardiomediastinal silhouette are normal in size and contour & 0.119 & 0.476 & 3.0 & 0.413 \\
report699 & 2 & opacity on the lateral view over the heart, present on previous exam, suggesting chronic subsegmental atelectasis or scarring & osseous structures are intact & 0.100 & 0.586 & 3.0 & 0.359 \\
report699 & 3 & no definite pleural effusion seen & there is no focal air space pleural or pneumothorax & 0.239 & 0.448 & 1.0 & 0.346 \\
report1860 & 1 & no focal airspace opacity & apparent interval increase in low density at the left \texttt{<unk>} & 0.050 & 0.550 & 3.0 & 0.176 \\
report1860 & 2 & lungs are clear & apparent interval increase in low density at the left \texttt{<unk>} & 0.067 & 0.400 & 2.0 & 0.160 \\
report1860 & 3 & levocurvature of the lumbar spine with significant degenerative change & apparent interval increase in low density at the left \texttt{<unk>} & 0.033 & 0.584 & 4.0 & 0.151 \\
\bottomrule
\end{tabular}}
\end{table*}

\begin{table}[tbp]
\centering
\scriptsize
\caption{Representative feature contributions from qualitative cases. Contributions are approximate scaled monotone nonnegative least-squares (NNLS) terms.}
\label{tab:contrib_app}
\begin{tabular}{llcc}
\toprule
Case/target & Feature & Coeff. & Contribution \\
\midrule
report699 total & Modifier expected & 7.224 & 4.770 \\
report699 total & Comparison top-3 MW & 3.534 & 3.534 \\
report699 total & Effective edges & 19.877 & 3.019 \\
report699 sig. & Modifier expected & 6.824 & 4.505 \\
report699 sig. & Comparison top-3 MW & 3.404 & 3.404 \\
report1860 insig. & Reference unit count & 3.972 & 3.404 \\
report1860 insig. & Anatomy expected & 1.496 & 0.980 \\
report1860 insig. & Effective edges & 2.630 & 0.920 \\
\bottomrule
\end{tabular}
\end{table}

\bibliographystyle{IEEEtran}
\bibliography{references}

\end{document}